\documentclass[sigconf]{acmart}

\usepackage{times}
\usepackage{listings}
\usepackage{xcolor}
\usepackage{color}
\usepackage{latexsym}
\usepackage{graphicx}
\usepackage{amsmath}
\usepackage{amsthm}
\usepackage{booktabs}
\usepackage{algorithm}
\usepackage{algorithmic}
\usepackage{caption}
\usepackage{subfigure}
\usepackage{multirow}
\usepackage{times}
\usepackage{amsfonts}

\usepackage{microtype}
\usepackage{footmisc}

\AtBeginDocument{%
  \providecommand\BibTeX{{%
    \normalfont B\kern-0.5em{\scshape i\kern-0.25em b}\kern-0.8em\TeX}}}

\copyrightyear{2021} 
\acmYear{2021} 
\setcopyright{acmcopyright}\acmConference[CIKM '21]{Proceedings of the 30th ACM International Conference on Information and Knowledge Management}{November 1--5, 2021}{Virtual Event, QLD, Australia}
\acmBooktitle{Proceedings of the 30th ACM International Conference on Information and Knowledge Management (CIKM '21), November 1--5, 2021, Virtual Event, QLD, Australia}
\acmPrice{15.00}
\acmDOI{10.1145/3459637.3482236}
\acmISBN{978-1-4503-8446-9/21/11}

\begin{document}
\fancyhead{}

\title{Behind the Scenes: An Exploration of Trigger Biases \\ Problem in Few-Shot Event Classification }

% \author{Peiyi Wang$^{\spadesuit}$\quad Runxin Xu$^{\spadesuit}$
% \quad Tianyu Liu \quad Damai Dai \quad Baobao Chang\quad Zhifang Sui
% }
% \affiliation{%
%   \institution{Key Laboratory of Computational Linguistics, Peking University, MOE, China}
%  }
% \email{{wangpeiyi9979, runxinxu}@gmail.com, {tianyu0421, daidamai, chbb, szf}@pku.edu.cn}
% % \email{{tianyu0421, daidamai, chbb, szf}@pku.edu.cn}
% \affiliation{$^{\spadesuit}$ Equal Contribution.}

\author[Peiyi Wang, Runxin Xu, Tianyu Liu, Damai Dai, Baobao Chang, and Zhifang Sui]{Peiyi Wang$^*$, Runxin Xu$^*$, Tianyu Liu, Damai Dai, Baobao Chang, and Zhifang Sui}

\makeatletter
\def\authornotetext#1{
 \g@addto@macro\@authornotes{%
 \stepcounter{footnote}\footnotetext{#1}}%
}
\makeatother

\authornotetext{Equal contributions. Order decided by tossed coins.}

\affiliation{%
 \country{Key Laboratory of Computational Linguistics, Peking University, MOE, China}
}
\email{{wangpeiyi9979, runxinxu}@gmail.com, {tianyu0421, daidamai, chbb, szf}@pku.edu.cn}

\def\authors{Peiyi Wang, Runxin Xu, Tianyu Liu, Damai Dai, Baobao Chang, and Zhifang Sui}

\renewcommand{\shortauthors}{Trovato and Tobin, et al.}

\begin{abstract}
  Few-Shot Event Classification (FSEC) aims at developing a model for event prediction, which can generalize to new event types with a limited number of annotated data. Existing FSEC studies have achieved high accuracy on different benchmarks.
However, we find they suffer from trigger biases that signify the statistical homogeneity between some trigger words and target event types, which we summarize as \emph{trigger overlapping} and \emph{trigger separability}. The biases can result in \emph{context-bypassing} problem, i.e., correct classifications can be gained by looking at only the trigger words while ignoring the entire context. Therefore, existing models can be weak in generalizing to unseen data in real scenarios.
To further uncover the trigger biases and assess the generalization ability of the models, we propose two new sampling methods, Trigger-Uniform Sampling (TUS) and COnfusion Sampling (COS), for the meta tasks construction during evaluation.
Besides, to cope with the context-bypassing problem in FSEC models, we introduce adversarial training and trigger reconstruction techniques. Experiments show these techniques help not only improve the performance, but also enhance the generalization ability of models. Our data and code is available at: \url{https://github.com/Wangpeiyi9979/Behind-the-Scenes}.

% Few-Shot Event Classification (FSEC) aims at developing a model for event prediction, which can generalize to new event types with a limited number of annotated data. Existing FSEC studies have achieved high accuracy on different benchmarks. However, we find they suffer from trigger biases that signify the statistical homogeneity between some trigger words and target event types, which we summarize as trigger overlapping and trigger separability. The biases can result in context-bypassing problem, i.e., correct classifications can be gained by looking at only the trigger words while ignoring the entire context. Therefore, existing models can be weak in generalizing to unseen data in real scenarios. To further uncover the trigger biases and assess the generalization ability of the models, we propose two new sampling methods, Trigger-Uniform Sampling (TUS) and COnfusion Sampling (COS), for the meta tasks construction during evaluation. Besides, to cope with the context-bypassing problem in FSEC models, we introduce adversarial training and trigger reconstruction techniques. Experiments show these techniques help not only improve the performance, but also enhance the generalization ability of models.
\end{abstract}





\maketitle

\section{Introduction}
\begin{figure}[tbp]
    \centering
    \includegraphics[width=1.0\linewidth]{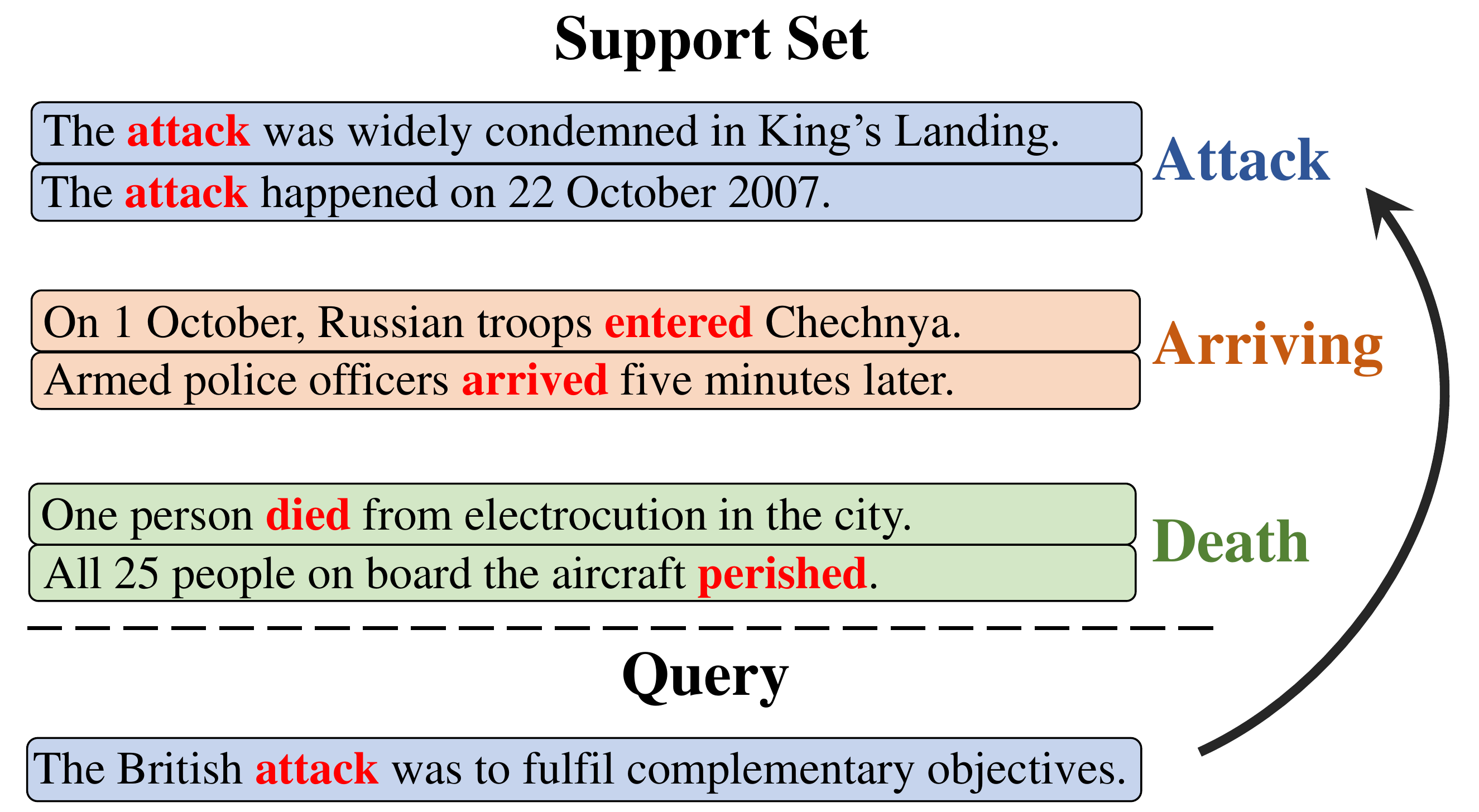}
    \caption{A meta task of $3$-way-$2$-shot Few-Shot Event Classification selected from MAVEN.
    There are $3$ event types (\textit{Attack}, \textit{Arriving}, and \textit{Death}) with $2$ instances in support set.
    The query instance is predicted to express the \textit{Attack} event type. 
    Note that all triggers of \textit{Attack} instances are \textit{attack}, which we call trigger overlapping and discussed in Section~\ref{sec:trigger-bias}. }
    \label{fig:running-example}
\end{figure}

Event Classification (EC) is an important task in Information Extraction (IE).
It aims at identifying specific types of events expressed in the text, and
the event is usually signaled by a trigger, i.e., the word that evokes the event.
Most traditional studies follow the supervised learning paradigm, which requires large-scale annotated data and is also limited to predefined event types.
To ease the burden of data annotation and develop event classification models that can generalize to new event types, few-shot learning has been introduced to event classification, i.e., Few-Shot Event Classification (FSEC).
FSEC usually adopts the meta-learning framework, which consists of a series of meta tasks.
For each meta task, given event types with their instances in the support set, we need to predict which event type the query instance belongs to.
Figure~\ref{fig:running-example} illustrates a $3$-way-$2$-shot meta task  , (i.e., $3$ event types with $2$ instances for each type), containing \textit{Attack}, \textit{Arriving}, and \textit{Death} events with their triggers colored in red.
The query instance is predicted as an \textit{Attack} event.

Most existing FSEC studies are based on Prototypical Network~\citep{NIPS2017_cb8da676} and achieve promising performance.
Prototypical Network makes predictions according to the semantic distance between prototype vectors for event types and query instance embeddings.
\cite{DBLP:conf/pakdd/LaiDN20} and \cite{lai-etal-2020-extensively} further propose auxiliary losses, and \cite{10.1145/3336191.3371796} introduces a dynamic memory module into the prototypical network.
These models all perform well on different datasets like MAVEN~\citep{wang-etal-2020-maven},  FewEvent~\citep{10.1145/3336191.3371796}, and ACE05~\citep{ace05}.
For example, \cite{DBLP:conf/pakdd/LaiDN20} can already achieve up to $87.2$ accuracy on ACE05 under $5$-way-$10$-shot setting.

However, \textbf{are existing FSEC models really generalize well to unseen data in real scenarios?}
In this paper, we investigate this issue in depth.
As we noticed, there exist unbalanced distributions in current datasets, and existing studies construct meta tasks by uniform sampling from all instances in such datasets.
We find it would bring about severe trigger biases, which we summarize as \emph{trigger overlapping} and \emph{trigger separability}.
These trigger biases can further lead to \emph{context-bypassing} problem in FSEC models, i.e., the model only relies on triggers to make predictions and totally ignores the context.
Context-bypassing problem would make the model overly rely on spurious \emph{trigger-event} alignment pattern in the dataset, and thus unable to generalize to the unseen data.
Take the following $2$-way-$1$-shot task as an example: 
\begin{table}[h]
\raggedright
% \vspace{-0.5em}
    \resizebox{8.0cm}{0.95cm}{
        \begin{tabular}{p{0.1cm}ll}
        %\multicolumn{2}{c}{} \\
        % \hline
        \textbf{$\mathcal{S}$}:&In 2011, Steve Jobs \textbf{left} this world. & [Death] \\
        ~ & President \textbf{went} to London to start a visit. &[Move] \\
        $q_1$: & The flight \textbf{left} Washington last night. &[Move] \\
        $q_2$: & Grandpa \textbf{went} to heaven. &[Death] \\
        % \hline
        %\multicolumn{2}{c}{} \\
        \end{tabular}
    }
% \vspace{-0.8em}
\end{table}

% \begin{tabular}{|p{1cm}|p{2cm}|p{3cm}|}
\noindent
where $\mathcal{S}$ is support set, $q_1$, $q_2$ are two queries, and $[\cdot]$ denotes event type.
Over-relying on the trigger overlapping pattern would mistakenly cause the model to choose the wrong event type.
Furthermore, \textit{left} and \textit{went} are semantically similar, hence the model can only make correct choices if context is effectively modeled.

To further uncover the trigger biases in the data and better assess the generalization ability, we design two new sampling methods for meta tasks construction, Trigger-Uniform Sampling (TUS) and COnfusing Sampling (COS), to intentionally remove the bias in the data.
Our experiments show that the accuracy of existing methods dramatically decreases by $20\%\sim35\%$ on data generated by TUS and COS sampling methods.

To cope with the context-bypassing problem caused by the trigger biases, we introduce two techniques, adversarial training and trigger reconstruction.
Experiments show they not only help improve the performance of the event classification model, but also enhance its generalization ability towards unseen data.

In summary, our contributions are three-fold.
1) To our best knowledge, we are the first to point out the trigger biases on Few-Shot Event Classification (FSEC), which may hurt the generalization ability of the classification models.
2) We propose two new sampling methods for meta tasks construction to assess the generalization ability of FSEC models.
3) We introduce two simple yet effective techniques, adversarial training and trigger reconstruction, to cope with the context-bypassing problem.

\section{Background}
In this section, we introduce the task formulation of Few-Shot Event Classification (FSEC) 
% (Sec.~\ref{sec:task-formulation})
and the sampling method for meta tasks construction.
%  (Sec.~\ref{sec:IUS})
We also introduce Prototypical Network widely adopted by previous works in FSEC.
% (Sec.~\ref{sec:prototypical-network}).

\subsection{Task Formulation}
\label{sec:task-formulation}
Following~\citep{DBLP:conf/pakdd/LaiDN20}, a meta task $\mathcal{T}$ in FSEC is formulated as follows.
Under $N$-way-$K$-shot setting, the model is given a support set $\mathcal{S}$ with $N$ event types, and there are $K$ instances for each event type, 
\begin{align*}
\mathcal{S} = &  \{ (s_1^1, p_1^1, e_1), \dots, (s_1^K, p_1^K, e_1), \\
& \dots \\
& (s_N^1, p_N^1, e_N), \dots, (s_N^K, p_N^K, e_N) \} 
\end{align*}
where $(s_i^j, p_i^j, e_i)$ denotes that event type $e_i$ is expressed by sentence $s_i^j$ and the $p_i^j$-th word in $s_i^j$ is the trigger.
Based on the support set $\mathcal{S}$, the goal is to predict the event type of the query $q=\left (s_q, p_q \right)$, where $s_q$ is the query sentence and $p_q$ indicates the trigger position.
Thus, we denote a meta task as $\mathcal{T}=\left (\mathcal{S}, q \right)$.
Both the training and test dataset consist of a series of meta tasks constructed by sampling from the datasets, $\mathcal{D}_\mathrm{train}=\left \{ \mathcal{T}^{(i)} \right \}_{i=1}^{\left | \mathcal{D}_\mathrm{train}  \right |}$, $\mathcal{D}_\mathrm{test}=\left \{ \mathcal{T}^{(i)} \right \}_{i=1}^{\left | \mathcal{D}_\mathrm{test}  \right |}$, and their label space are guaranteed to be disjoint with each other.

\subsection{Instance-Uniform Sampling}
\label{sec:IUS}
Previous studies construct FSEC meta tasks (i.e., support set and query) by Instance-Uniform Sampling (IUS) from event classification datasets.
In detail, under $N$-way-$K$-shot settings, IUS firstly uniformly sample $N$ different event types.
Then, for each event type, IUS uniformly sample $K$ instances from all instances of this event type to form the support set.
The query construction is similar, i.e., randomly choose one of the $N$ event types and then uniformly sample from all according instances.

\subsection{Prototypical Network}
\label{sec:prototypical-network}
Most previous methods~\citep{DBLP:conf/pakdd/LaiDN20,lai-etal-2020-extensively,10.1145/3336191.3371796,DBLP:journals/corr/abs-2012-02353} are based on Prototypical Network~\citep{NIPS2017_cb8da676}.
Prototypical Network calculates a prototype vector, $c_k \in \mathbb{R}^{d_{m}}$  for each event type through a neural network $f_\theta$ with trainable parameters $\theta$:
\begin{equation*}
    c_k = \frac{1}{\left | \mathcal{S}_k \right | } \sum_{(s_k^i,p_k^i,e_k) \in \mathcal{S}_k} f_\theta (s_k^i, p_k^i)
\end{equation*}
where $\mathcal{S}_k$ denotes the set of instances labeled with event type $e_k$, and $s^i_k$, $p^i_k$ are the corresponding sentence and trigger position.
Given a distance function $D:  \mathbb{R}^{d_m} \times \mathbb{R}^{d_m} \longrightarrow \mathbb{R} $, we predict the query instance $q=(s_q, p_q)$ as the $e_k$ event type in the support set with probability:
\begin{equation*}
    P(y=e_k|s_q, p_q) = \frac{exp(-D(f_\theta(s_q,p_q), c_k)}{\sum_{k'} exp(-D(f_\theta(s_q,p_q), c_{k'}))} 
\end{equation*}

\section{Trigger Bias Problem}

In this section, we discuss the unbalanced distribution on current benchmarks (Sec.~\ref{sec:dataset}).
We then demonstrate that the unbalanced distribution, along with instance-uniform sampling for meta tasks construction, results in trigger overlapping and trigger separability biases (Sec.~\ref{sec:trigger-bias}).
Finally, we introduce the context-bypassing  problem caused by these trigger biases (Sec.~\ref{sec:context-bypassing}).

\subsection{Unbalanced Distribution of Datasets}
\label{sec:dataset}

Currently, meta-learning usually construct meta tasks by sampling data from annotated datasets.
For event classification, there are three popular datasets, MAVEN~\citep{wang-etal-2020-maven},  FewEvent~\citep{10.1145/3336191.3371796}, and ACE05~\citep{ace05}.
FewEvent is the extension version of ACE05, with more event types and instances.
We dive into these datasets and find that there are two noticeable distribution patterns.

\begin{figure}[t]
    \centering
    \includegraphics[width=1.0\linewidth]{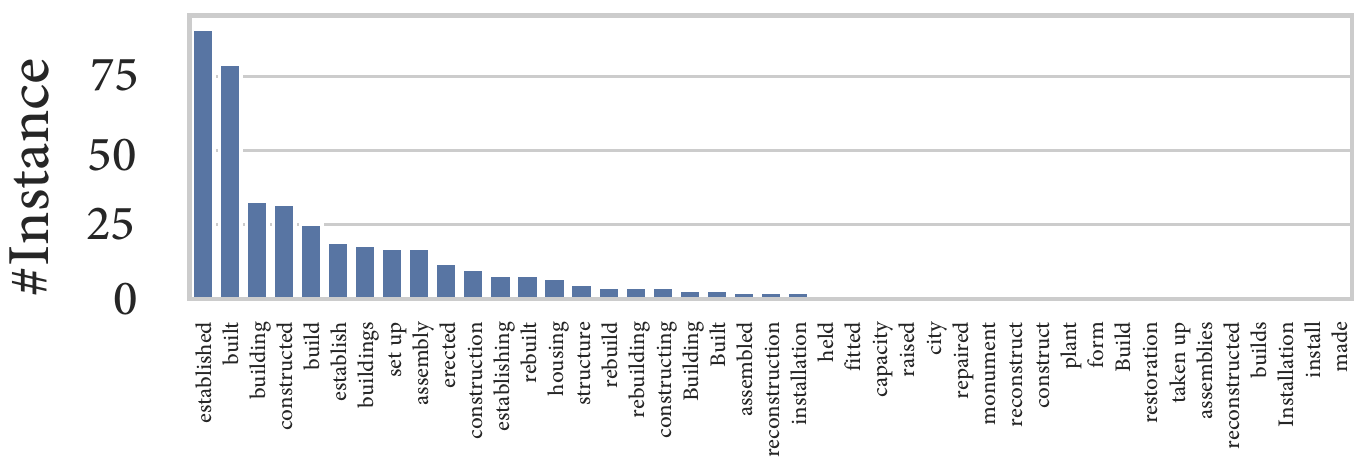}
    \caption{Long-tail distribution of triggers of ``\textit{Building}'' event type in MAVEN dataset.}
    \label{fig:long-tail}
\end{figure}

Firstly, \emph{the long-tail distribution of triggers for an event}. 
Figure~\ref{fig:long-tail} shows an example of the \textit{Building} event type in MAVEN.
A majority of instances contain top-frequent triggers, such as \textit{established}.
Statistically, as shown in Table~\ref{table:trigger-long-tail}, there are more than $60\%$ instances involved with the top-$5$ frequent triggers of an event type on average, although an event type usually contains far more than $5$ triggers, e.g., $66$ triggers on average in MAVEN.
Therefore, most instances of an event type are triggered by a small number of frequent triggers, while there are still some instances with many other triggers.

\begin{figure}[t]
    \centering
    \includegraphics[width=1.0\linewidth]{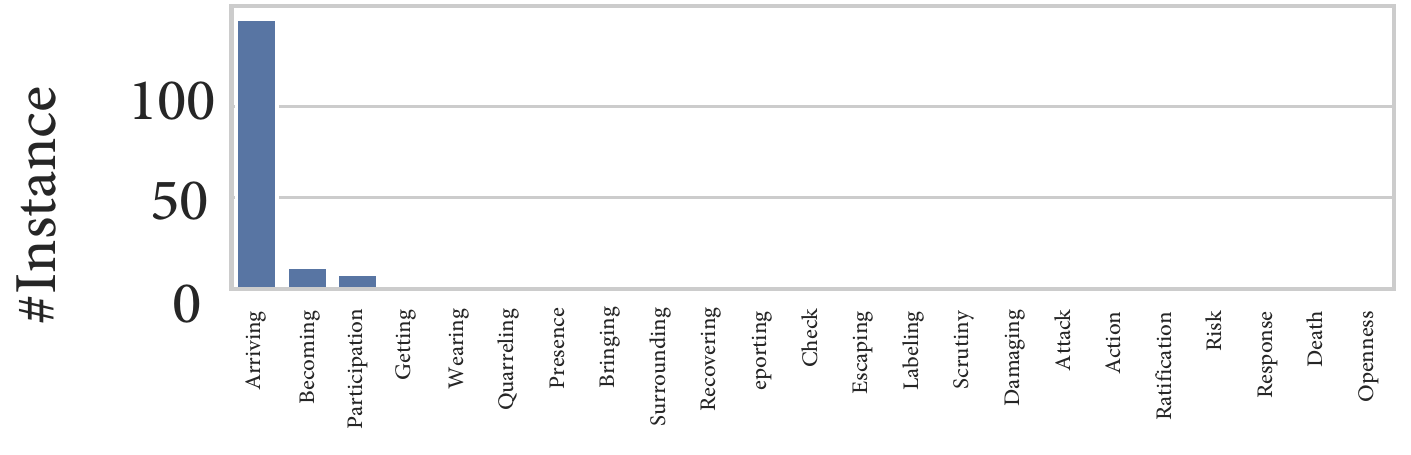}
    \caption{Skewed event types distribution for top-frequent trigger, ``\textit{enter}'', in MAVEN dataset.}
    \label{fig:skewed}
\end{figure}

Secondly, \emph{the skewed event type distribution of top-frequent triggers}.
Figure~\ref{fig:skewed} demonstrates an example of the top-frequent trigger of \textit{Arriving} event type, \textit{enter}.
Most instances with trigger \textit{enter} belongs to the \textit{Arriving} event type, and there is only $12.4\%$ of them belonging to other event types.
Table~\ref{table:event-long-tail} shows instances with top-$5$ triggers of different event types usually belong to only $1\sim3$ event types.
Besides, for a top-frequent trigger, more than $95\%$ instances with it belong to the top-$2$ dominant event types of this trigger on average.
It suggests that top-frequent triggers are usually strongly tied with their belonging dominant event types.

\subsection{Trigger Biases: Trigger Overlapping and Trigger Separability}
\label{sec:trigger-bias}

In this section, we introduce trigger overlapping and trigger separability biases, which are caused by the unbalanced distribution in datasets and the IUS method for meta tasks construction.

\begin{table}[t]
    \centering
    \caption{The statistics of $3$ datasets \textbf{for event types}. \texttt{\#Triggers}: average number of triggers for an event type.  \texttt{Top-5 Ins\%}: average proportion of instances with top-$5$ frequent triggers of an event type.
    }
    \begin{tabular}{lccc}
    \toprule
    \bf Dataset & \bf \#Event Types & \bf \#Triggers & \bf Top-5 Ins\%\\
    \midrule
    MAVEN & 168 & 66 & 63\% \\
    FewEvent & 100 & 42 & 68\% \\
    ACE05 & 33 & 50 & 60\% \\
    \bottomrule
    \end{tabular}
    \label{table:trigger-long-tail}
\end{table}

\begin{table}[t]
    \centering
    % \small
    \caption{The statistics of $3$ datasets \textbf{for top-5 frequent triggers}. \texttt{\#Avg Events}: the average number of event types that a top-$5$ frequent trigger belongs to. \texttt{Top-x Ins\%}: among all instances with a certain top-$5$ trigger, the average proportion of instances (with this trigger) that belong to the top-x dominant event types.
    }
    \begin{tabular}{lccc}
    \toprule
    \bf Dataset & \bf \#Avg Events & \bf Top-1 Ins\% & \bf Top-2 Ins\%\\
    \midrule
    MAVEN & 3.10 & 78\% & 95\% \\
    FewEvent & 2.38 & 75\% & 96\% \\
    ACE05 & 1.48 & 90\% & 99\% \\
    \bottomrule
    \end{tabular}
    \label{table:event-long-tail}
\end{table}

\begin{figure}[t]
\centering     
\subfigure[GloVe embeddings of top-$5$ frequent triggers.]{  
\begin{minipage}{7cm}
\centering  
\includegraphics[width=1.0\linewidth]{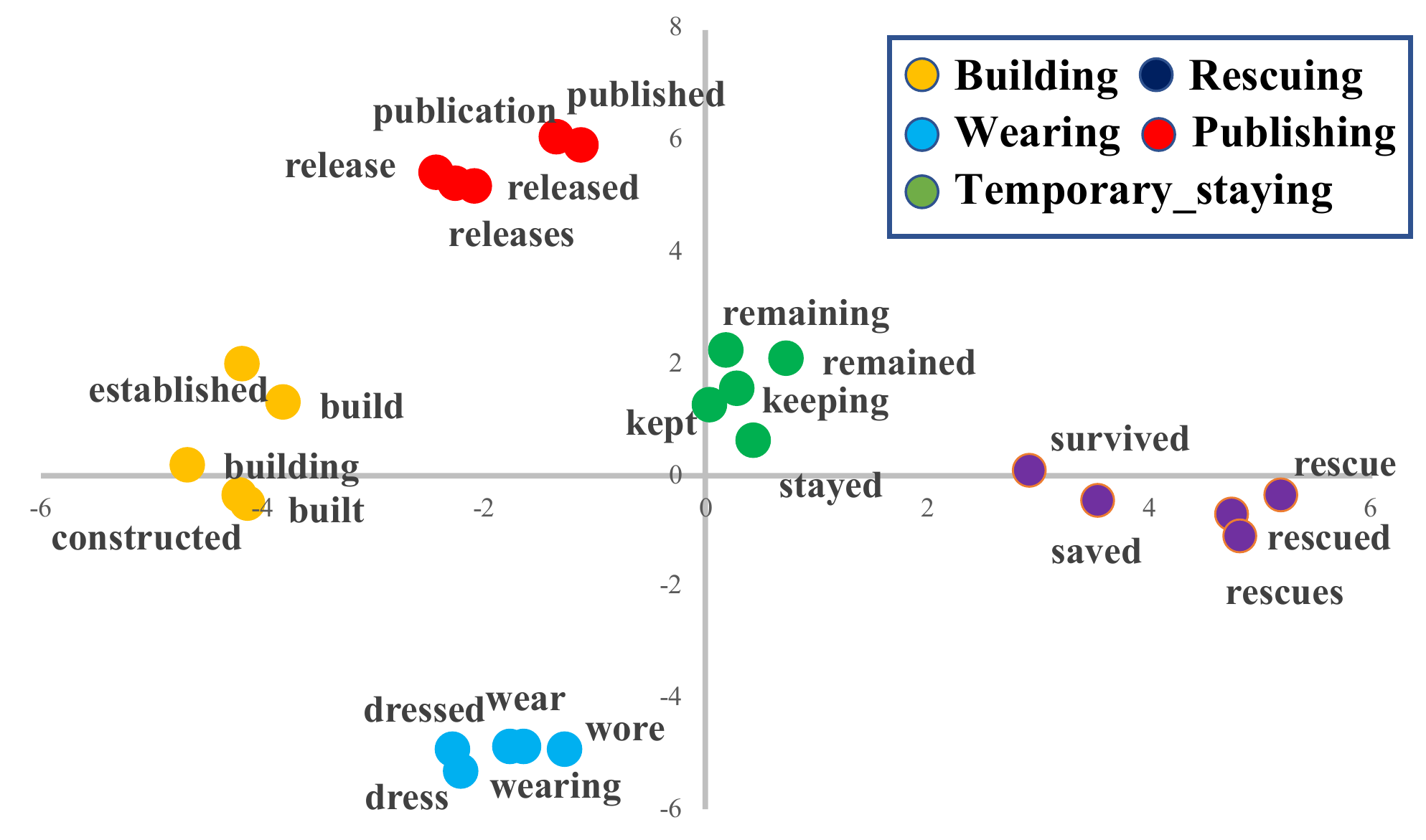}              
\end{minipage}}
\subfigure[GloVe embeddings of all triggers.]{
\begin{minipage}{7cm}
\centering 
\includegraphics[width=1.0\linewidth]{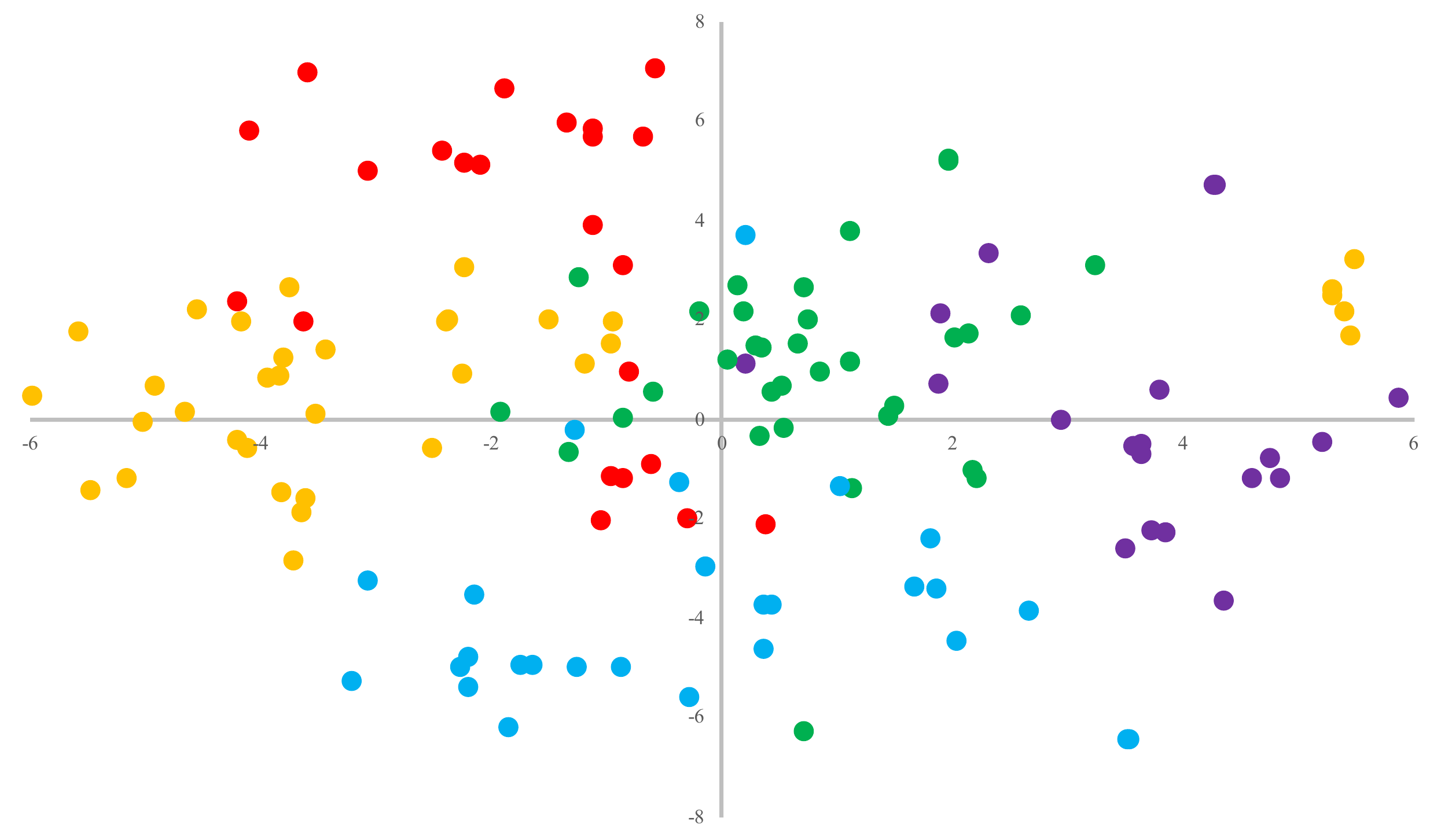}            
\end{minipage}}
\caption{GloVe embeddings of triggers of $5$ event types in MAVEN dataset using t-SNE~\citep{maaten2008visualizing}. Triggers can be easily separated if only top-$5$ frequent triggers are considered, while it becomes chaotic with all triggers considered.}
\label{fig:trigger-separable}
\end{figure}

\begin{table*}
\caption{Accuracy under $4$ different $N$-way-$K$-shot settings. We report the average accuracy of $5$ random trials, along with the standard deviation. Results above the double line do not use GloVe embedding. Though ignoring all the context, String Match and GloVe Match still achieve comparable performance in comparison with other neural contextualized model.}
\centering
 
\subtable[\textbf{Results in FewEvent Dataset.}]{
\begin{tabular}{lcccc}
\toprule
\bf Model & \bf 5-way-5-shot & 
\bf 5-way-10-shot & \bf 10-way-5-shot & 
\bf 10-way-10-shot \\
\midrule
String Match & 68.51 $\pm$ 0.13 & 77.29 $\pm$ 0.12 & 64.47 $\pm$ 0.05 & 74.37 $\pm$ 0.14 \\
Proto-CNN w/o GloVe Emb & 61.21 $\pm$ 1.15 & 69.86 $\pm$ 1.15 & 50.31 $\pm$ 2.00 & 57.13 $\pm$ 2.50  \\
Proto-BiLSTM w/o GloVe Emb & 63.12 $\pm$ 0.38 & 70.49 $\pm$ 0.47 & 57.39 $\pm$ 0.36 & 64.94 $\pm$ 0.71  \\
\midrule
\midrule
GloVe Match & 84.90 $\pm$ 0.14 & 87.57 $\pm$ 0.13 & 79.10 $\pm$ 0.01 & 83.02 $\pm$ 0.05 \\
Proto-CNN  &  81.09 $\pm$ 1.52 & 84.21 $\pm$ 0.39 & 71.95 $\pm$ 0.65 & 77.80 $\pm$ 0.62  \\
Proto-BiLSTM &  86.86 $\pm$ 0.18 & 89.06 $\pm$ 0.15 & 80.06 $\pm$ 0.33 & 84.09 $\pm$ 0.27  \\
\bottomrule
\end{tabular}
\label{table:problem-result-fewevent}
}

\subtable[\textbf{Results in MAVEN Dataset.}]{        
\begin{tabular}{lcccc}
\toprule
\bf Model & \bf 5-way-5-shot & 
\bf 5-way-10-shot & \bf 10-way-5-shot & 
\bf 10-way-10-shot \\
\midrule
String Match & 61.06 $\pm$ 0.19 & 60.94 $\pm$ 0.11 & 32.42 $\pm$ 0.14 & 67.19 $\pm$ 0.12 \\
Proto-CNN w/o GloVe Emb &  41.88 $\pm$ 0.39 & 49.64 $\pm$ 1.40 & 31.18 $\pm$ 1.03 & 38.34 $\pm$ 1.93  \\
Proto-BiLSTM w/o GloVe Emb & 47.37 $\pm$ 0.66 & 57.02 $\pm$ 0.70 & 39.86 $\pm$ 0.41 & 49.54 $\pm$ 0.53  \\
\midrule
\midrule
GloVe Match & 84.96 $\pm$ 0.09 & 88.62 $\pm$ 0.14 & 78.22 $\pm$ 0.13 & 82.67 $\pm$ 0.09 \\
Proto-CNN  &  82.44 $\pm$ 0.86 & 86.49 $\pm$ 0.55 & 74.92 $\pm$ 0.74 & 78.87 $\pm$ 0.32  \\
Proto-BiLSTM &  85.61 $\pm$ 0.89 & 88.65 $\pm$ 0.80 & 78.95 $\pm$ 0.41 & 82.80 $\pm$ 0.43  \\

\bottomrule
\end{tabular}
\label{table:problem-result}
}
\end{table*}

\textbf{Trigger Overlapping}
\quad Since instances with top-frequent triggers are more likely to be sampled by IUS, triggers in the meta task are limited to top-frequent triggers with high probability.
Hence, the trigger of the query is very likely to be identical to triggers of some instances in the support set, which we call trigger overlapping bias.
As shown in Figure~\ref{fig:running-example}, the trigger of both the query and instances of \textit{Attack} event type are \textit{attack}.
Hence, the model can correctly predict through this bias without considering the context.
For further illustration, we randomly construct $102,000$ meta tasks under $5$-way-$5$-shot from the MAVEN and FewEvent dataset.
We find there are $55,611$ and $54,503$ meta tasks having such trigger overlapping bias,  respectively.
% The number will increase to $**$ and $**$ if we use $10$-way-$10$-shot setting.
Trigger overlapping bias dramatically weaken the generalization ability because 
1) the overlapping may mislead the prediction of the classification models,
2) and the context around is ignored while it is informative and helpful.
% and there are new emergent triggers in practice which do not overlap with triggers of existing instances.

\textbf{Trigger Separability}
\quad With IUS, top-frequent triggers are sampled with much more probability, and these triggers are usually tied with their belonging dominant event types.
Therefore, instances are easily separable in feature space by looking at only the triggers.
Figure~\ref{fig:trigger-separable} illustrates an example, in which we randomly choose $5$ event types (\textit{Building}, \textit{Rescuing}, \textit{Temporary\_staying}, \textit{Wearing}, and \textit{Publishing}) in MAVEN.
We show the GloVe embedding of their top-$5$ triggers in Figure~\ref{fig:trigger-separable} (a), where they are close to triggers in the same event type and far away from those in different event types. 
However, as shown in Figure~\ref{fig:trigger-separable} (b), when it comes to all triggers in the $5$ event types, the separability pattern disappears and triggers distribution become chaotic in feature space.
Hence, if the model over-rely on the shallow features of triggers to separate instances, it can have difficulty handling triggers that do not bind with specific event types.

\citep{shah-etal-2020-predictive} has defined four potential origins of biases: \emph{label bias}, \emph{selection bias}, \emph{model overamplification}, and \emph{semantic bias}.
The two proposed trigger biases belong to selection bias and model overamplification, in that
1) the selected data rarely consider a majority of less frequent triggers and more challenging cases, 
2) and the model tends to predict based on wrong clues that cannot generalize to unseen data in real scenarios.
% , such as trigger overlapping.

\subsection{Context-Bypassing Problem}
\label{sec:context-bypassing}

Trigger overlapping and trigger separability biases bring the serious context-bypassing problem, i.e., the over-dependency on triggers in the sentence and the ignorance of useful information in the context.
Previous FSEC models suffer from the context-bypassing problem.
Besides, previous evaluation methods also have difficulty reflecting the generalization ability of models, and therefore the performance in previous studies may be overestimated.

We hypothesize that a model can still achieve high accuracy even if it only considers the trigger.
To verify our hypothesis, we propose two extremely simple methods, i.e., String Match and GloVe Match, which \textbf{totally abandon the contextual information of the instance}.
For String Match, we choose the event type containing the overlapped triggers with the query as our prediction, if the trigger overlapping occurs.
Otherwise, we just randomly select an event type as our prediction.
For Glove Match, we adopt the Prototypical Network, but only utilize the GloVe embedding of the triggers as the instance representations, without any neural architecture or surrounding contextual information.
We compare them with CNN/BiLSTM-based Prototypical Network~\citep{10.1145/3336191.3371796}, which use CNN/BiLSTM as the encoder to encode the whole sentence to obtain instance representation, whose experimental settings are introduced in Section~\ref{sec:experiments-setting}. 

We conduct experiments on FewEvent and MAVEN dataset.
As shown in Table.~\ref{table:problem-result-fewevent} and Table~\ref{table:problem-result},
String Match outperforms CNN/BiLSTM-based model without GloVe embedding in most cases. For example, String Match achieves $11.53$ and $15.14$ accuracy improvement on average in FewEvent and MAVEN dataset, in comparison with CNN-based model.
In addition, GloVe Match surprisingly achieves comparable performance with the BiLSTM-based model, and even beat the CNN-based model with $4.89$ average accuracy improvement in FewEvent dataset, and $2.94$ in MAVEN dataset.
These results support our claims that current evaluation methods cannot distinguish whether the model truly comprehends the semantic information and can well generalize to unseen data, or they just simply utilize the trigger biases without any comprehension.

\section{TUS and COS sampling methods}
\label{sec:sampling}

To further uncover the trigger biases and assess the generalization ability of models, we propose two sampling methods, trigger-uniform sampling and confusion sampling to construct meta task.
Similar to IUS, we uniformly sample $N$  event types, but the following procedures are different.

\subsection{Trigger-Uniform Sampling}
Trigger-Uniform Sampling (TUS) is proposed to ease the trigger overlapping bias.
When constructing meta tasks, for event type $e$ in support set $\mathcal{S}$, we uniformly sample $K$ triggers from all triggers of $e$, \emph{regardless of their occurring frequency}. 
Then we uniformly sample an instance from all instances of $e$ with the corresponding trigger.
For query $q$, we randomly choose one of the $N$ event types, and the other operations are the same.

Treating each trigger equally, TUS ensures the triggers between query and support set are usually not overlapped, and the less frequent triggers are also considered.
We expect that the evaluation can prevent the models from taking advantage of trigger overlapping bias to achieve high accuracy.

\subsection{Confusion Sampling}
COnfusion Sampling (COS) is proposed to ease the trigger separability bias, which pays more attention to confusing triggers. 
In detail, COS consists of two steps: trigger partition and trigger sampling.

\textbf{Trigger partition}
\quad Considering a meta task $\mathcal{T}$, for each event type $e$, we split all of its triggers $T_e$ into confusing $S_{con}$ and non-confusing sets $S$.
Intuitively, a trigger $t$ is confusing for an event type $e$, if $t$ is relatively different from other triggers in the same event type $T_e$, and relatively similar to triggers in other event types $T_{e_o}$ in $\mathcal{T}$.
For example, some triggers may be shared across different events, and they can be confusing for their belonging events, such as trigger \textit{offer} for event type \textit{Financing} and \textit{Employment}.
The detailed partition process is shown in Algorithm ~\ref{alg:cos2}, where we use $L2$ distance of GloVe embedding to measure the similarity of triggers, and pick up confusing triggers according to the comprehensive distance $d_{com}$ we define in Line $6$.
Note that the confusion set $S_{con}$ for an event type $e$ would vary in different meta tasks, as the other event types $e_o$ would  change.

\textbf{Trigger sampling}
\quad We sample trigger $t$ for each event type $e$ from confusion set $S_{con}$ with a controlling probability $p$, otherwise from non-confusion set $S$.
Then we uniformly sample an instance from all instances with trigger $t$ of the corresponding event type.
The operation for query is similar.

In general, instances with confusing triggers can be correctly predicted only if the context is utilized, because the superficial features of triggers may be misleading.
Therefore, confusing sampling can avoid the models to make use of trigger separability bias to gain high accuracy during evaluation.

~\\
\indent Note that when we change IUS to TUS or COS to construct meta tasks during evaluation, we find the accuracy of existing models dramatically decreases by $17.09\sim29.17$ (Sec.~\ref{sec:sampling-results}), which suggests that the trigger biases do exist and they bring about serious context-bypassing problem.

\renewcommand{\algorithmicrequire}{ \textbf{Input:}} %Use Input in the format of Algorithm
\renewcommand{\algorithmicensure}{ \textbf{Output:}} %UseOutput in the format of Algorithm

\begin{algorithm}[t] 
% \small
\caption{Trigger Partition for an Event Type} 
\label{alg:cos2}
\begin{algorithmic}[1]
\REQUIRE  %算法的输入参数：Input
$e$: Event type whose triggers are to be partitioned. \\
$O$: Other events in the task. \\
$T_{e}$: Triggers set of event $e$. \\
$E_{t}$: GloVe embedding of trigger $t$. \\
$D$: L2 distance function. \\
$U$: Hyperparameters controlling the size of $S_{con}$.
\ENSURE 
% ~~\\ %算法的输出：Output
Confusing set $S_{con}$ and non-confusing set $S$.
\STATE $S_{con} \gets \emptyset$
\FOR{all $e_o \in O$}
    \FOR{all $t \in T_e$}
        \STATE {$d_{inner}^t = \frac{1}{\left \| T_e \right \| }  \sum_{t' \in T_e}D(E_t, E_{t'})$}
        \STATE $d_{inter}^t = \frac{1}{\left \| T_{e_o} \right \| }  \sum_{t' \in T_{e_o}}D(E_t, E_{t'})$
        \STATE $d_{com}^t = -d_{inner}^t+d_{inter}^t$
    \ENDFOR
    \STATE {Add top-$U$ triggers with smallest $d_{com}$ into $S_{con}$.}
\ENDFOR
\STATE $S \gets T_e - S_{con}$
\RETURN $\left \{S_{con}, S \right \}$
\end{algorithmic}
\end{algorithm}

\section{Strategies to Handle Context- Bypassing Problem}
In this section, we introduce adversarial training and trigger reconstruction strategies to handle the context-bypassing problem in FESC models.
Both of them try to mitigate the over-reliance on the trigger to improve the generalization ability of models.

\subsection{Adversarial Training}
The adversarial training method is introduced by \citep{DBLP:journals/corr/GoodfellowSS14}.
Specifically, given sample $X$ and its label $y$, adversarial training tries to add some noise $\delta$, where $\left \|  \delta \right \| \le \epsilon$ and $\epsilon$ is a constant, such that the loss function $\mathcal{L}(\cdot,\cdot)$ is maximized by $\delta$.
Therefore, the final loss function of adversarial training to optimize model parameters $\theta$ is the following min-max objective:
\begin{equation}
    \min_{\theta} \mathbb{E}_{(X, y)\sim \mathcal{D}}\left [  \max_{ \left \|  \delta \right \| \le \epsilon } 
\mathcal{L}(f_\theta(X+\delta),y) \right ] 
\end{equation}

In our paper, we add such noise to the trigger embedding, which tries to cut off the statistical homogeneity between triggers and event types and hence enforce the model to consider the context.
Specifically, we train the model in the following way.
Firstly, we calculate the cross entropy loss $\mathcal{L}_{ce}$ and derive the gradients as normal.
Secondly, according to~\citep{DBLP:conf/iclr/MiyatoDG17}, we add $\delta = \frac{\epsilon g_{tri}  }{\left \| g_{tri} \right \|_2 } $ to the trigger embedding, where $g_{tri}$ denotes the gradients of trigger embedding.
The motivation is that the gradient is the direction of the steepest ascent for the loss function, and therefore we add noise in this direction for the adversarial attack.
Thirdly, we calculate the loss $\mathcal{L}_{adv}$ another time with this new noisy trigger embedding.
Finally, we sum up the two losses with weight $\alpha$ as follows:
\begin{align}
\mathcal{L} = \mathcal{L}_{ce} + \alpha\mathcal{L}_{adv}
\end{align}
and update the parameters of the models, where $\alpha$ is an controlling hyperparameter.
In this way, the model is forced to also consider the context, and therefore enhance the generalization ability.

\subsection{Trigger Reconstruction}
We propose to reconstruct the trigger word through the context to enhance the ability of the model to comprehend the context, and therefore mitigate the context-bypassing problem.
In detail, we mask the trigger, i.e., replace the trigger with a special token like \texttt{[MASK]}, and reconstruct the trigger token based on all the other tokens, i.e., the context around.
In the implementation, we use the contextualized hidden state in the trigger position after encoding to predict the trigger token.
This is similar to the masked language modeling task~\citep{devlin2019bert}, while there are still some important differences.
a) We only reconstruct the trigger token rather than random tokens.
b) We mask the token with $100\%$ probability.
We denote the reconstruction loss as $\mathcal{L}_{rec}$ and derive the loss as follows with weight $\beta$:
\begin{align}
    \mathcal{L} = \mathcal{L}_{ce} + \beta\mathcal{L}_{rec}
\end{align}

\section{Experiments}

\begin{table*}
\caption{Accuracy under different sampling methods for meta tasks construction. We report the average accuracy of $5$ random trials. \textbf{Mean $\Delta$} denotes the mean accuracy difference under TUS/COS in comparison with IUS across $4$ $N$-way-$K$-shot settings. The accuracy of all models dramatically decreases when changing the sampling methods from IUS to TUS/COS.}
\centering

\subtable[\textbf{Results on FewEvent Dataset.}]{
\scalebox{0.95}{
    \begin{tabular}{l|cccccccccccc|cc}
    \toprule
    \multirow{2}{*}{\bf Model} & \multicolumn{3}{c}{\bf 5-way-5-shot} & 
    \multicolumn{3}{c}{\bf 5-way-10-shot} & 
    \multicolumn{3}{c}{\bf 10-way-5-shot} & 
    \multicolumn{3}{c|}{\bf 10-way-10-shot} &
    \multicolumn{2}{c}{\bf Mean $\Delta$} \\
    \cmidrule(lr){2-4} \cmidrule(lr){5-7} \cmidrule(lr){8-10} \cmidrule(lr){11-13}
    \cmidrule(lr){14-15}
    ~ & \bf IUS & \bf TUS & \bf COS & \bf IUS & \bf TUS & \bf COS & \bf IUS & \bf TUS & \bf COS & \bf IUS & \bf TUS & \bf COS & \bf TUS & \bf COS   \\
    \midrule
    String Match  & 68.51 & 19.51 & 19.36 & 77.29 & 19.13 & 18.97 & 64.47 & 9.46 & 9.38 & 74.37 & 9.12 & 8.97 & -56.86 & -56.99 \\
    GloVe Match &   84.90 & 57.88 & 54.54 & 87.57 & 61.81 & 60.70 & 79.10 & 46.15 & 43.27 & 83.02 & 50.26 & 48.21 & -29.62 & -31.97 \\
    \midrule
    % MN  &  00.00 & 00.00 & 00.00 & 00.00 & 00.00 & 00.00 & 00.00 & 00.00 & 00.00 & 00.00 & 00.00 & 00.00  & -00.00 & -00.00\\
    Proto-CNN  & 81.09 & 61.84 & 60.50 & 84.21 & 63.60 & 63.58 & 71.95 & 48.75 & 48.15 & 77.80 & 51.54 & 50.12 & -22.33 & -23.18 \\
    Proto-BiLSTM  &  86.86 & 63.43 & 59.70 & 89.06 & 66.70 & 65.55 & 80.06 & 50.12 & 46.49 & 84.09 & 54.11 & 51.67 & -26.43 & -29.17 \\
    HATT  &  82.95 & 64.30 & 59.76 & 85.24 & 66.33 & 64.75 & 75.24 & 51.94 & 47.35 & 79.47 & 55.09 & 52.75 & -21.31 & -24.57 \\
    HATT+LoLoss  &  83.40 & 64.50 & 62.54 & 87.35 & 67.04 & 66.82 & 76.88 & 51.33 & 49.80 & 80.47 & 55.17 & 54.89 & -22.52 & -23.51 \\
    HATT+Loss$_{\rm int}$  & 82.69 & 64.60 & 62.36 & 86.15 & 65.84 & 65.17 & 74.85 & 51.09 & 48.48 & 79.84 & 54.95 & 52.47 & -21.76 & -23.76 \\
    \midrule
    Proto-BERT  &  94.01 & 78.89 & 78.22 & 95.24 & 80.61 & 81.75 & 90.83 & 69.85 & 69.94 & 93.03 & 72.58 & 74.83 & -17.80 & -17.09 \\
    HATT-BERT  &  94.96 & 80.01 & 77.75 & 95.44 & 79.16 & 79.77 & 90.86 & 70.84 & 68.05 & 92.72 & 68.73 & 68.92 & -18.81 & -19.87 \\
    \bottomrule
    \end{tabular}
}
\label{table:fewevent-main}
}
 
\subtable[\textbf{Results on MAVEN Dataset.}]{        
\scalebox{0.95}{
    \begin{tabular}{l|cccccccccccc|cc}
    \toprule
    \multirow{2}{*}{\bf Model} & \multicolumn{3}{c}{\bf 5-way-5-shot} & 
    \multicolumn{3}{c}{\bf 5-way-10-shot} & 
    \multicolumn{3}{c}{\bf 10-way-5-shot} & 
    \multicolumn{3}{c|}{\bf 10-way-10-shot} &
    \multicolumn{2}{c}{\bf Mean $\Delta$} \\
    \cmidrule(lr){2-4} \cmidrule(lr){5-7} \cmidrule(lr){8-10} \cmidrule(lr){11-13}
    \cmidrule(lr){14-15}
    ~ & \bf IUS & \bf TUS & \bf COS & \bf IUS & \bf TUS & \bf COS & \bf IUS & \bf TUS & \bf COS & \bf IUS & \bf TUS & \bf COS & \bf TUS & \bf COS   \\
    \midrule
    String Match  &  61.06 & 19.82 & 19.59 & 71.85 & 19.65 & 19.19 & 55.36 & 9.82 & 9.69 & 67.19 & 9.63 & 9.39 & -49.14 & -49.40 \\
    GloVe Match &  84.96 & 59.71 & 45.70 & 88.62 & 66.65 & 52.44 & 78.22 & 45.8 & 35.03 & 82.67 & 52.68 & 42.35 & -27.41 & -39.74 \\
    \midrule
    % MN  &  00.00 & 00.00 & 00.00 & 00.00 & 00.00 & 00.00 & 00.00 & 00.00 & 00.00 & 00.00 & 00.00 & 00.00  & -00.00 & -00.00\\
    Proto-CNN  &  82.44 & 57.15 & 46.19 & 86.49 & 62.60 & 53.21 & 74.92 & 42.73 & 34.67 & 78.87 & 48.21 & 41.53 & -28.01 & -36.78 \\
    Proto-BiLSTM  & 85.61 & 60.95 & 49.04 & 88.65 & 67.43 & 56.54 & 78.95 & 46.43 & 37.09 & 82.80 & 53.72 & 44.62 & -26.87 & -37.18 \\
    HATT & 84.08 & 57.60 & 45.90 & 87.77 & 63.32 & 52.23 & 76.41 & 42.98 & 34.31 & 81.33 & 49.59 & 41.17 & -29.02 & -39.00 \\
    HATT+LoLoss  & 84.98 & 57.65 & 46.51 & 88.82 & 64.49 & 54.13 & 77.72 & 44.02 & 35.00 & 82.62 & 50.71 & 41.33 & -29.32 & -39.29 \\
    HATT+Loss$_{\rm int}$  & 83.75 & 57.02 & 46.47 & 87.93 & 64.06 & 52.40 & 76.10 & 42.58 & 34.23 & 81.55 & 49.00 & 40.81 & -29.17 & -38.86 \\
    \midrule
    Proto-BERT  & 90.48 & 69.31 & 60.46 & 92.37 & 75.30 & 68.06 & 84.89 & 55.91 & 47.81 & 87.95 & 62.65 & 56.38 & -23.13 & -30.75 \\
    HATT-BERT  &  90.07 & 65.01 & 55.56 & 92.94 & 71.73 & 61.38 & 85.17 & 51.94 & 43.14 & 88.09 & 58.95 & 49.15 & -27.16 & -36.76 \\
    \bottomrule
    \end{tabular}
}
\label{table:maven-main}
}
 
\end{table*}
\begin{table*}[htbp]
\centering
% \small
\caption{Accuracy on FewEvent. We report the average accuracy of $5$ random trials. Using adversarial training (+adv) and trigger reconstruction (+rec) leads to improvements under different sampling methods for evaluation.
}
\scalebox{1.0}{
    \begin{tabular}{l|cccccccccccc}
    \toprule
    \multirow{2}{*}{\bf Model} & \multicolumn{3}{c}{\bf 5-way-5-shot} & 
    \multicolumn{3}{c}{\bf 5-way-10-shot} & 
    \multicolumn{3}{c}{\bf 10-way-5-shot} & 
    \multicolumn{3}{c}{\bf 10-way-10-shot} \\
    \cmidrule(lr){2-4} \cmidrule(lr){5-7} \cmidrule(lr){8-10} \cmidrule(lr){11-13}
    ~ & \bf IUS & \bf TUS & \bf COS & \bf IUS & \bf TUS & \bf COS & \bf IUS & \bf TUS & \bf COS & \bf IUS & \bf TUS & \bf COS   \\
    \midrule
    Proto-CNN  & 81.09 & 61.84 & 60.50 & 84.21 & 63.60 & 63.58 & 71.95 & 48.75 & 48.15 & 77.80 & 51.54 & 50.12 \\
    +adv  & 81.30 & 65.49 & 63.42 & \bf 85.92 & 66.95 & 66.04 & 73.19 & 51.77 & 51.45 & 78.72 & 54.55 & \bf 54.59 \\
    +rec  & \bf 83.12 & 64.78 & 63.57 & 85.31 & 65.61 & 64.04 & 72.71 & 51.41 & 50.39 & 79.07 & 53.89 & 52.70 \\
    +adv+rec  & 82.49 & \bf 65.83 & \bf 65.09 & 85.83 & \bf 68.35 & \bf 67.40 & \bf 73.53 & \bf 52.84 & \bf 52.27 & \bf 79.66 & \bf 55.28 & 53.68 \\
    \midrule
    Proto-BERT   & 94.01 & 78.89 & 78.22 & 95.24 & 80.61 & 81.75 & 90.83 & 69.85 & 69.94 & 93.03 & 72.58 & 74.83 \\
    +adv  & 94.31 & 79.81 & 79.28 & 95.64 & 82.07 & 82.81 & 91.22 & \bf 71.91 & 70.86 & \bf 93.52 & 74.44 & 76.01 \\
    +rec  & 94.17 & 79.14 & 78.63 & 95.61 & 81.26 & 82.80 & 91.18 & 71.26 & 70.87 & 93.24 & 73.07 & 75.45 \\
    +adv+rec  & \bf 94.45 & \bf 79.87 & \bf 79.62 & \bf 95.75 & \bf 82.80 & \bf 83.18 & \bf 91.57 & 71.73 & \bf 72.30 & 93.25 & \bf 75.03 & \bf 76.35  \\
    \bottomrule
    \end{tabular}
}

\label{table:fewevent-adv-rec}
\end{table*}

\subsection{Dataset}
We conduct the experiments on two popular event classification dataset,
MAVEN~\citep{wang-etal-2020-maven} and FewEvent~\citep{10.1145/3336191.3371796}.
As FewEvent is the extension of ACE05~\citep{ace05}, we omit the experiments on ACE05.
MAVEN contains $51,173$ instances for $168$ event types.
We filter out event types containing less than $100$ instances, with $134$ event types remained.
FewEvent contains $67,841$ instances for $100$ event types.
We also filter event types containing less than $40$ instances and obtain the remaining $48$ event types.
we randomly choose $109$/$12$/$13$ and $27$/$10$/$11$  event types for train/dev/test set for MAVEN and FewEvent respectively, which involves $33,993$/$4,329$/$12,851$ and $49,269$/$7,171$/$11,401$ instances respectively.
% More details can be found in Appendix~\ref{sec:dataset-division}.

\subsection{Experimental  Setting}
\label{sec:experiments-setting}

% Our experiments settings are demonstrated in Table~\ref{tab:hyperparameters}.
Our experiments settings are as follows,
for CNN-based models, we set the window size and filter number to $3$ and $300$, and use max-pooling to obtain instance representation.
For BiLSTM-based models, we set the hidden size to $300$ with a single layer.
We use $300$d GloVe~\citep{glove} embedding, and $50$d position embedding which indicates the distance between current token to trigger token following \citep{DBLP:conf/pakdd/LaiDN20}.
For BERT~\citep{devlin2019bert}, we use the BERT-base-uncased version provided by Huggingface\footnote{https://huggingface.co/bert-base-uncased/tree/main}.
We use the final hidden states in trigger position as instance representations for BiLSTM/BERT-based model.
We set $\epsilon=0.5$ for adversarial training, $\alpha=1.0$ and $\beta=0.1$ for trigger reconstruction.
Adam~\cite{kingma2014adam} is used as optimizer, and the learning rate is set to $3e-5$ and $1e-4$ for BERT and non-BERT modules.
Besides, we set a patience number to $3$, so that we could stop the learning early if there is no further performance improvement on validation set, and choose the best checkpoint in validation set to evaluate on the test set.
We set $U=6$ and $p=1.0$ for COS.
We construct $10,000$ meta tasks for each evaluation, and report the average accuracy of $5$ different random seeds.
We use a single GeForce RTX 3090 GPU to run our experiments.

\subsection{Baselines}
We use the following models as baselines.
1) \textbf{Proto-CNN}/\textbf{Proto-BiLSTM}/\textbf{Proto-BERT}: \citep{NIPS2017_cb8da676} proposes Prototypical Network as introduced in Section.~\ref{sec:prototypical-network} and they use $L2$ distance. 
In our paper, we use CNN, BiLSTM, and BERT as the encoder for Prototypical Network respectively.
2)  \textbf{HATT/HATT-BERT}: \citep{Gao_Han_Liu_Sun_2019} proposes HATT based on Prototypical Network, which uses hybrid attention, i.e., instance-level attention to focus on query-related instances, and feature-level attention to pay more attention to discriminative feature dimensions. We use both CNN encoder (HATT) and BERT encoder (HATT-BERT).
3) \textbf{HATT+LoLoss}: On top of HATT, ~\citep{DBLP:conf/pakdd/LaiDN20} introduces an auxiliary loss, LoLOSS, which is calculated by taking some instances of support set as queries to construct extra meta tasks. It can also be viewed as a method of data augmentation. We use CNN as encoder.
4) \textbf{HATT+Loss$_{\mathrm{int}}$}: ~\citep{lai-etal-2020-extensively} proposes two auxiliary loss, Loss$_{\mathrm{intra}}$ to minimize the distance between instances in the same class, and Loss$_{\mathrm{inter}}$ to maximize distance between pairs of prototypes. We use Loss$_{\mathrm{int}}$ to refer to both, and CNN as encoder.

\subsection{Results Under New Sampling Methods}
\label{sec:sampling-results}
We have proposed TUS and COS for meta tasks construction (Section.~\ref{sec:sampling}).
In this section, we reveal the experimental results of different models when using these two new sampling methods, which aims to:
1) prove the trigger biases and context-bypass problems do exist,
2) and illustrate the strength of our proposed sampling methods to assess the generalization ability of models.

As Table~\ref{table:fewevent-main} shown, for the FewEvent dataset, although all the methods achieve promising results under IUS, their performance dramatically decrease if the sampling methods are changed to TUS and COS.
Specifically, since the trigger biases are eased by TUS and COS, methods that ignore the context, e.g., String Match, drops $56.86$ accuracy under TUS in comparison with IUS, and even drops $56.99$ accuracy under COS (indicated by \textbf{Mean $\Delta$}).
Besides, non-BERT models also have a decline of accuracy ranging from $21.31\sim26.43$ and $23.18\sim29.17$ under TUS and COS, suggesting that these models may over-rely on trigger biases and the generalization ability is damaged.
The degree of accuracy decline is relatively small for models with powerful contextualized encoder BERT, while it is still up to $19.87$ accuracy decrease.
The experimental results in MAVEN dataset is similar, which is shown in Table~\ref{table:maven-main}.
Adopting TUS method leads to $23.13\sim29.32$ accuracy decrease for different models, and it is even worse when we adopt COS method to construct meta tasks, with up to $30.75\sim39.29$ accuracy decrease.

In summary, changing traditional IUS method into TUS or COS to construct meta tasks leads to significant performance decrease. 
It suggests that the event classification models are overestimated since they take advantage of the trigger biases, and using TUS and COS can help assess the generalization ability of these models.

\begin{table*}[h]
\centering
% \small
\caption{Accuracy on MAVEN. We report the average accuracy of $5$ random trials. Using adversarial training (+adv) and trigger reconstruction (+rec) leads to improvements under different sampling methods for evaluation.
}
\scalebox{1.0}{
    \begin{tabular}{l|cccccccccccc}
    \toprule
    \multirow{2}{*}{\bf Model} & \multicolumn{3}{c}{\bf 5-way-5-shot} & 
    \multicolumn{3}{c}{\bf 5-way-10-shot} & 
    \multicolumn{3}{c}{\bf 10-way-5-shot} & 
    \multicolumn{3}{c}{\bf 10-way-10-shot} \\
    \cmidrule(lr){2-4} \cmidrule(lr){5-7} \cmidrule(lr){8-10} \cmidrule(lr){11-13}
    ~ & \bf IUS & \bf TUS & \bf COS & \bf IUS & \bf TUS & \bf COS & \bf IUS & \bf TUS & \bf COS & \bf IUS & \bf TUS & \bf COS   \\
    \midrule
     Proto-CNN  &   82.44 & 57.15 & 46.19 & 86.49 & 62.60 & 53.21 & 74.92 & 42.73 & 34.67 & 78.87 & 48.21 & 41.53 \\
    +adv & 83.39 & 58.54 & 48.05 & \bf 87.63 & 64.44 & \bf 55.44 & 76.22 & 44.09 & 35.55 & \bf 80.58 & 50.86 & 42.83 \\
    +rec  & 83.35 & 56.94 & 47.19 & 87.03 & 63.37 & 53.63 & 75.36 & 43.41 & 35.23 & 79.90 & 49.04 & 42.07 \\
    +adv+rec  & \bf 83.90 & \bf 58.93 & \bf 48.26 & 87.58 & \bf 64.96 & 55.21 & \bf 76.27 & \bf 44.34 & \bf 36.12 & 80.30 & \bf 51.06 & \bf 43.06 \\
    \midrule
    Proto-BERT & 90.48 & 69.31 & 60.46 & 92.37 & 75.30 & 68.06 & 84.89 & 55.91 & 47.81 & 87.95 & 62.65 & 56.38 \\
    +adv  & 90.67 & 69.89 & 60.68 & 92.74 & 75.43 & 69.71 & 85.93 & 56.89 & 49.07 & 88.11 & \bf 64.12 & 57.70 \\
    +rec  & 90.59 & 69.70 & 60.51 & 92.24 & 75.43 & 68.59 & 85.32 & 56.84 & 48.71 & 88.01 & 63.90 & 57.48 \\
    +adv+rec  & \bf 90.80 & \bf 70.70 & \bf 61.30 & \bf 92.82 & \bf 76.03 & \bf 70.06 & \bf 86.00 & \bf 57.18 & \bf 49.44 & \bf 88.34 & 64.11 & \bf 57.72 \\
    \bottomrule
    \end{tabular}
}
\label{table:maven-adv-rec}
\end{table*}

\begin{figure*}[t]
    \centering
    \includegraphics[width=1\linewidth]{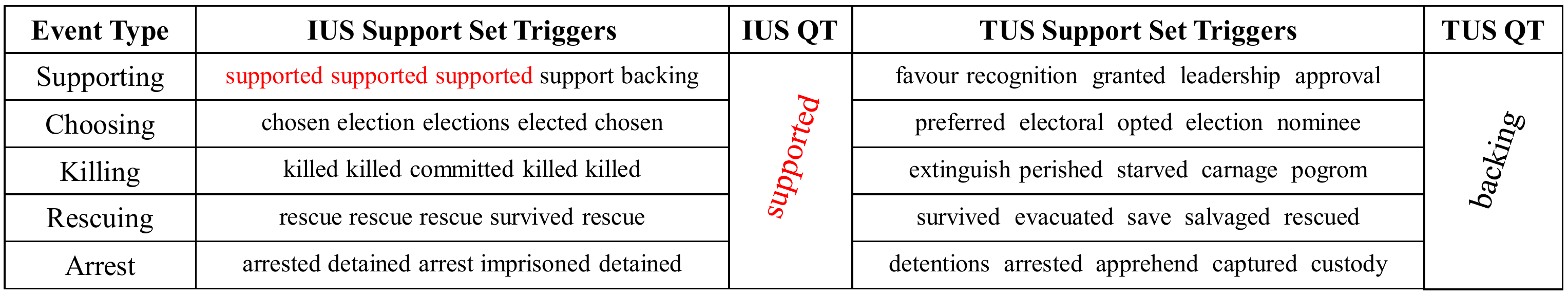}
    \caption{Comparison of triggers of two meta tasks constructed by Instance Uniform Sampling (IUS) and Trigger Uniform Sampling (TUS). The task constructed by IUS has serious trigger overlapping bias (\emph{supported} is the trigger of both query and support set instances), while TUS eases this bias.
    QT: Query Trigger.}
    \label{fig:IUSTUS}
    % \vspace{0.2cm}
\end{figure*}

\begin{figure}[t]
    \centering
    \includegraphics[width=1.0\linewidth]{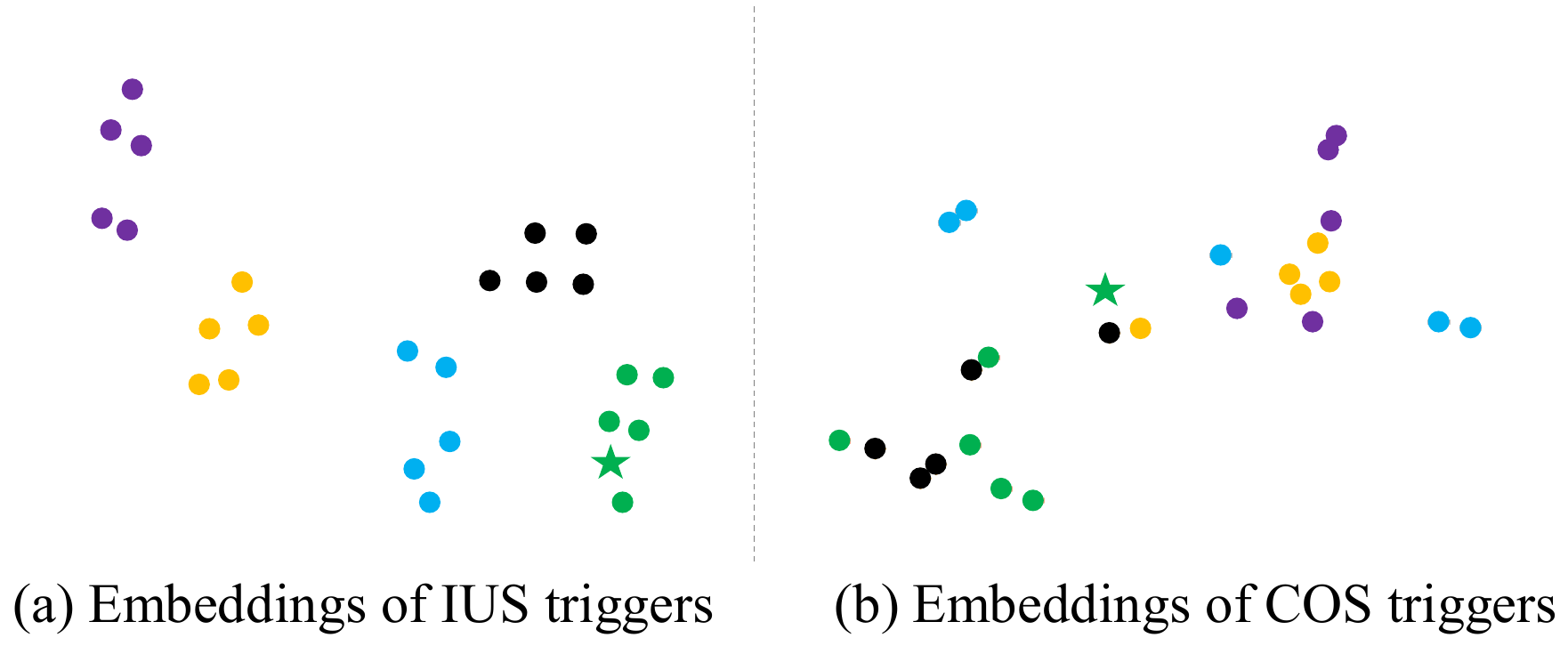}
    \caption{Comparison of trigger GloVe embeddings of meta tasks constructed by (a) Instance Uniform Sampling (IUS) method and  (b) COnfusion Sampling (COS) method, using t-SNE.
    The circulars represent triggers in the support set, and the pentacle represents trigger of the query. Different colors represent different event types.}
    \label{fig:IUSCOS}
\end{figure}

\subsection{Results with Proposed Strategies}
As shown in Table~\ref{table:fewevent-adv-rec} and Table~\ref{table:maven-adv-rec}, using adversarial training (+adv) and trigger reconstruction (+rec) yield significant improvements across different $N$-way-$K$-shot settings on both two datasets.
Take $10$-way-$5$-shot for example.
When we use traditional IUS method to construct meta tasks, adversarial training (+adv) and trigger reconstruction (+rec) lead to $1.24$ and $0.76$ score improvement for Proto-CNN on FewEvent, while combing both of them can further achieves $1.58$ score improvement.
More importantly, when we use TUS or COS methods to construct meta tasks, both adversarial training and trigger reconstruction can also improve the performance of the model.
Specifically, for Proto-CNN model, adversarial training (+adv) and trigger reconstruction (+rec) yield an improvement of up to $3.65$/$4.47$ and $2.94$/$3.07$ for TUS/COS evaluation on FewEvent dataset, respectively, and $2.65$/$2.23$ and $0.83$/$1.00$ on MAVEN dataset.
Further integrating both of them (+adv+rec) can even achieves an improvement of up to $4.75$/$4.59$ and $2.85$/$2.07$ for TUS/COS evaluation on FewEvent and MAVEN dataset, respectively.

In summary, both the adversarial training and trigger reconstruction techniques can not only improve accuracy under traditional IUS evaluation, but also enhance the generalization ability of models and therefore achieve higher accuracy under TUS and COS evaluation.
Besides, the effect of adversarial training and trigger reconstruction are orthogonal to each other, and hence integrating both of them can further achieve higher performance boost.

\begin{figure*}[t]
    \centering
    \includegraphics[width=0.9\linewidth]{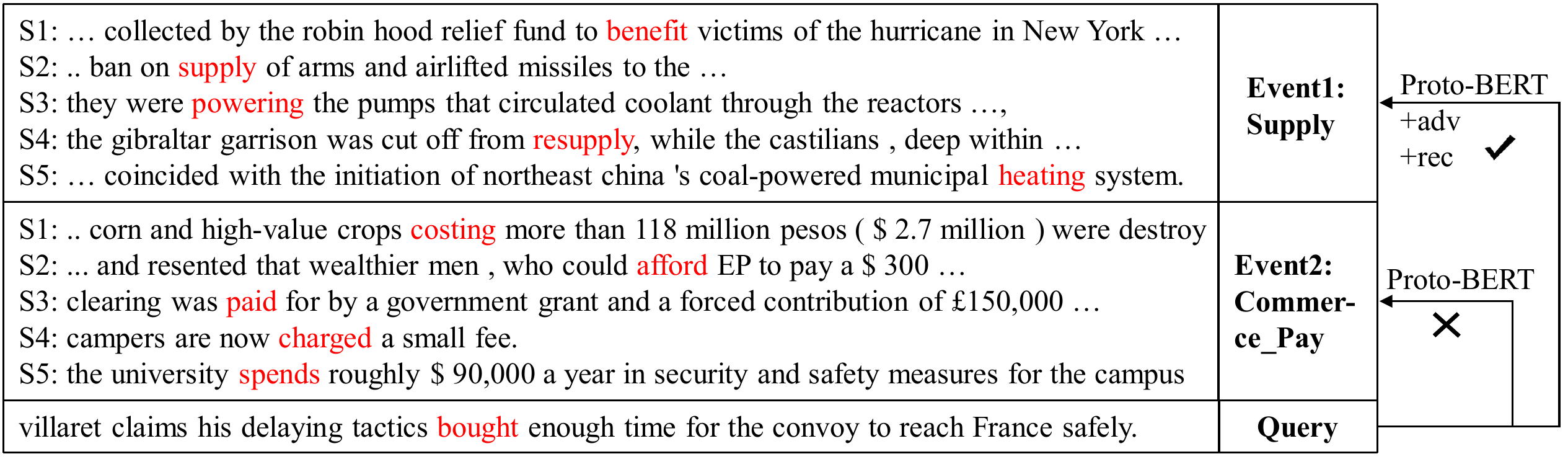}
    \caption{
    % A meta task that can not be solved by utilizing trigger biases.
    The red tokens represent triggers. Vanilla Proto-BERT has difficulty dealing with this meta task, since simply taking advantage of trigger biases would result in wrong prediction.
    Instead, training model with our proposed techniques encourage the model to mitigate the context-bypassing problem and therefore correctly predict.
    We only list related two event types.}
    \label{fig:adv-tr-train}
\end{figure*}

\section{Case Study}

\subsection{Effect of Sampling Methods}

\noindent\textbf{Effect of TUS}
\quad Figure \ref{fig:IUSTUS} illustrates two groups of triggers ($5$-way-$5$-shot setting) that are constructed by traditional IUS and our proposed TUS methods, respectively.
In detail, for the meta task constructed by traditional IUS method, the word \emph{supported} is the trigger of both query instance and support set instances, which exposes serious trigger overlapping bias and the model can address such meta task using this superficial features.
However, for the meta task constructed by our proposed TUS method, the trigger overlapping bias is alleviated since the triggers rarely overlap with each other.

\noindent\textbf{Effect of COS}
\quad Figure \ref{fig:IUSCOS} visualizes the GloVe embeddings of triggers in the meta tasks constructed by IUS and COS respectively. 
As shown in Figure~\ref{fig:IUSCOS} (a), the triggers of instances sampled by IUS is separable and therefore exposes trigger separability bias.
However, as shown in Figure~\ref{fig:IUSCOS} (b), constructing meta tasks through COS method encourages the triggers not to cluster, and hence eases the trigger separability bias.

\subsection{Effect of Training Strategies}
In order to achieve more robust model for FSEC, we design two training strategies, adversarial training and trigger reconstruction.
As shown in Figure \ref{fig:adv-tr-train},
the query instance with trigger \emph{bought} belongs to \emph{Supply} event type. 
However, vanilla Proto-BERT model has difficulty dealing with such meta task, 
because the trigger \emph{bought} is often used in commerce scenarios, and it is semantically closer to triggers of event \emph{Commerce\_Pay}.
Instead, training with adversarial training and trigger reconstruction techniques encourages the model to pay attention to the context, and therefore the model succeed to mitigate the context-bypassing problem and correctly predict.

\section{Related Work}

\textbf{Event Detection}
\quad Event detection is an important task in Information Extraction~\citep{zeng-etal-2020-double, xu-etal-2021-document}, which consists of Trigger Identification and Event Classification.
Trigger Identification aims at extracting all the event triggers from the sentence and Event Classification needs to classify them into the corresponding event type.
The previous  event detection methods can be broadly divided into two categories.
One is the two-stage model, which first performs Trigger Identification, then does Event Classification. 
For example, in the first stage,
\cite{ghaeini2016event} identifies triggers through string matching with training corpus and database. \cite{feng2016language} utilizes a bi-directional LSTM to extract triggers.
\cite{lin2018nugget} proposes a Trigger Nugget Generator to generate event triggers. 
Then, they propose different neural networks to do event classification with triggers and sentence as inputs.
The other is the one-stage model, which regards all the tokens in the sentence as trigger candidates, and performs event classification with an extra event type \textit{NA}, which means the corresponding token is not a trigger.
For example,
\cite{liu2017exploiting} proposes a neural network with supervised attention mechanisms.
\cite{nguyen2018graph} utilizes dependency tree and Graph Convolutional Network. 
\cite{liu2019exploiting} designs a teacher-student framework.
In this paper, we focus on Event Classification task, since Trigger Identification can be solved by the pre-stage model.

\noindent\textbf{Few-Shot Event Classification}
\quad To mitigate the data-hungry problem and generalize to new event types, Few-Shot Event Classification (FSEC) is proposed.
Most methods of FSEC adopt the meta-learning framework formulating FSEC as a sequence of tasks containing support sets and query instance.
On top of Prototypical Network~\cite{NIPS2017_cb8da676}, 
\cite{DBLP:conf/pakdd/LaiDN20} propose an auxiliary loss from the viewpoint of data augmentation.
\cite{lai-etal-2020-extensively} also introduce two new auxiliary losses for intra-event and inter-event regularization.
\cite{10.1145/3336191.3371796} further use dynamic memory module to extract richer semantic features.
\cite{DBLP:journals/corr/abs-2012-02353} formulate the this problem as a sequence-tagging task instead and introduces amortized conditional random field.
All these methods achieve high accuracy, however, we find that they are overestimated due to some trigger biases.

\noindent\textbf{Biases in NLP}
\quad \cite{shah-etal-2020-predictive} propose a unified framework to analyze biases in NLP, and define four potential origins of biases: \emph{label bias}, \emph{selection bias}, \emph{model overamplification bias}, and \emph{semantic bias}.
The trigger bias proposed in our paper belongs to \emph{selection bias} and \emph{model overamplification bias}.
Bias has also been investigated in natural language inference ~\citep{gururangan2018annotation, mccoy-etal-2019-right, belinkov-etal-2019-dont, feng2019misleading,liu-etal-2020-hyponli,liu-etal-2020-empirical,ding-etal-2020-discriminatively}, question answering~\citep{min-etal-2019-compositional}, ROC story cloze~\citep{DBLP:conf/acl/CaiTG17, DBLP:conf/conll/SchwartzSKZCS17}, lexical inference \cite{levy2015supervised}, visual question answering \cite{goyal2017making}, etc.
To our best knowledge, we are the first to present the biases in FSEC, i.e., trigger overlapping and trigger separability.

\section{Conclusion}
Despite the promising performance, previous Few-Shot Event Classification (FSEC) models may be overestimated due to trigger biases.
We analyze the causes of the trigger biases, and show that it can lead to the serious \emph{context-bypassing} problem, i.e., correct predictions can be gained by only the trigger without any context.
To further uncover the trigger biases problem and better assess the generalization ability of models, we propose two new sampling methods for more challenging meta tasks construction. 
Besides, we introduce adversarial training and trigger reconstruction techniques to handle the context-bypassing problem in FSEC models.

Our analysis can also extend to other few-shot learning tasks that construct meta tasks by simple uniform sampling from unbalanced data,
which we leave as future work.

\begin{acks}
The authors would like to thank the anonymous reviewers for their thoughtful and constructive comments.
This paper is supported by the National Key Research and Development Program of China 2020AAA0106700 and NSFC project U19A2065, and the National Science Foundation of China under Grant No. 61876004.
\end{acks}

\bibliographystyle{ACM-Reference-Format}
\bibliography{fewec}

% \appendix
% \input{main/appendix}

\end{document}